# Temporal Computer Organization

**J. E. Smith**

**University of Wisconsin-Madison (Emeritus)**

**January 19, 2022**

## Abstract

This document is focused on computing systems implemented in technologies that communicate and compute with temporal transients. Although described in general terms, implementations of spiking neural networks are of primary interest. As background, an algebra for constructing temporal networks is summarized. Then, a system organization consisting of synchronized segments is described. The segments are feedforward internally with feedback between segments. A synchronizing clock resets network segments at the end of each computation step or cycle. In its basic form, the synchronizing clock merely performs a reset function. In the context of neural networks, this satisfies biological plausibility. However, functional completeness is restricted. This restriction is removed by allowing use of the synchronizing clock as an additional function input that acts as a temporal reference value.

## 1. Introduction

A temporal computing system operates on values encoded as discrete temporal events: transients. In contrast to most digital technologies which are designed to signal via levels, some technologies are inherently transient in nature. The transients may be voltage or current pulses ("spikes") or changes in voltage or current levels ("edges"). Biological neural networks which communicate via action potentials are an example, and the primary motivation for this work is a silicon-implemented version: spiking neural networks [7][11][18]. However, there are other examples: superconducting technologies that communicate via single flux quantum pulses [6][16]. Any technology that relies on communication and computation via transient temporal events may benefit from the theory and methods described here.

*Note:* In this document, to simplify discussions, all types of transient events are generically referred to as "spikes".

The model developed here adheres to three key principles:

1) Communication is spike-based, and all computation takes spike encoded values as inputs and produces spike encoded outputs in a way that is consistent with the flow of physical time.

2) Communication and computation use very low precision unary encoded values. The underlying mathematics operates on small integer values. Very low precision is consistent with the properties and constraints of biological neural networks. Other transient-based technologies may be similarly constrained.

3) A synchronization mechanism divides computation into steps. Synchronization in biological neural networks [3] is consistent with Abeles synfire chains [1]. See Figure 1.

This document is focused on computing systems that satisfy the above three principles using technologies that communicate and compute with temporal transients. An effort is made to be as general as possible with spiking neural networks being an important case. As background, an algebra for constructing temporal networks is summarized in the next section. Then, Section 3 describes a system organization consisting of synchronized segments that are feedforward internally, with feedback between segments. A synchronizing signal, the *gamma clock*, patterned after biological gamma rhythms [3], resets network segments at the end of each cycle. In its basic form, the gamma clock is not used as a temporal reference value for computation -- it merely performs a reset function. In the context of neural networks, this satisfies biological plausibility, however, functional completeness is restricted. This restriction is removed in Section 4 by allowing use of the gamma clock as an additional function input that acts as a temporal reference value. Section 5 discusses the two clock synchronization method. One clock models the biological gamma clock, and the other is a unit time clock that is part of a silicon implementation.

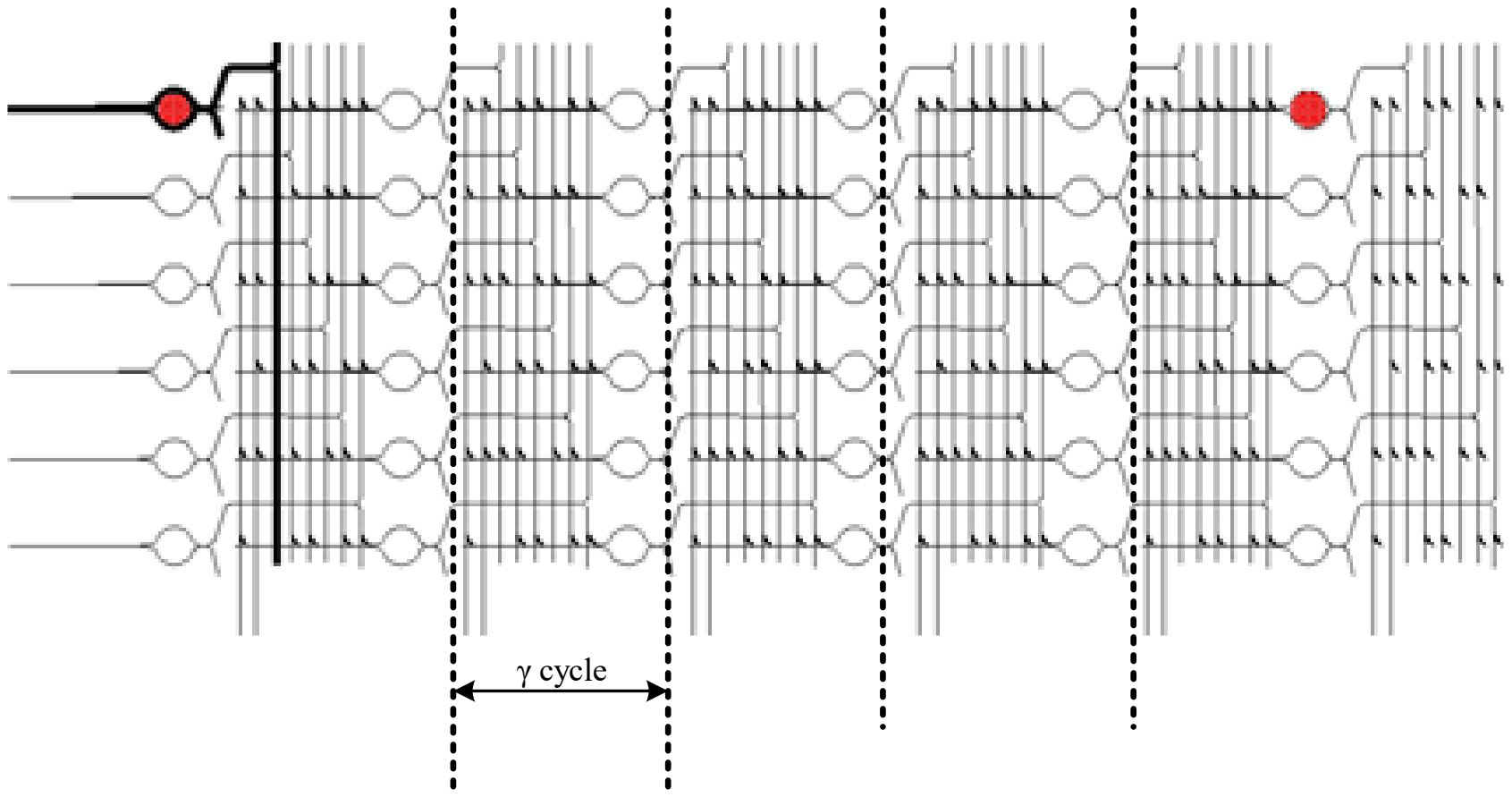

**Figure 1. An Abeles synfire chain organizes neurons (circles) into layers, separated by synaptic crossbars (small triangles). Spike volleys pass from one layer to the next, synchronized via gamma cycle inhibition. Each volley contains at most one spike per line. As pointed out by Abeles, the layout in the figure does not represent the anatomy, rather it represents the processing time sequence. Hence, the figure is drawn in an unrolled form for illustrative purposes; in reality, there can be any amount of feedback among layers. For example, the red neuron shown at two different points in the chain is the same physical neuron. (Figure taken from Scholarpedia [1] with timing notations added.)**

## 2. (Newtonian) Space-Time Algebra

For the construction of temporal computing systems, the role of *s-t* algebra [12] is analogous to the role of Boolean algebra in conventional digital hardware. The *s-t algebra* consists of 0, the natural numbers, and the symbol "∞" , i.e. $N_0^\infty$. The interpretation used here is that the values in the algebra represent the times of events. If an event occurs, it is given a value (time) that is relative to the occurrence of other events; if an event never occurs, it is assigned the symbol "∞".

### 2.1 Fundamentals

The algebra includes *min* and *max* operators: $\wedge$ and $\vee$ . The algebra is distributive, associative, commutative, and satisfies the absorption laws. However, the algebra is not complemented. Elements of the algebra may be added, but there is no subtraction because the algebra is closed only for addition, and all elements of the algebra are non-negative. Observe that when the algebra is given a temporal interpretation then subtraction (or complementation or negation) would be tantamount to going backwards in time. Hence, such operations are not supported in a system where operations are consistent with the forward flow of time.

**Definition:** A function $z = F(x_1...x_q)$, $x_1...x_q$ , $z \in N_0^\infty$, is a *space-time function* if it satisfies the following:

1) *implementability: q* is finite, and F is implementable with a finite number of states.

2) *causality*: i) For all $x_j > z$, $F(x_1...,x_j,...x_q) = F(x_1...,\infty,...x_q)$, and ii) if $z \neq \infty$, then $z \geq min(x_i)$ .

3) *invariance*: $F(x_1 + 1, ..., x_q + 1) = F(x_1,..x_q) + 1$. □

The three properties are very general. *Implementability* restricts consideration to functions that satisfy Church-Turing computability constraints. The other two properties are consistent with the uniform passage of time. *Causality*: using the temporal interpretation, an output spike can not be affected by input spikes that occur later in time. Furthermore, there are no spontaneous output spikes. *Invariance*: if all the input spikes uniformly shift by unit time, then the output spike shifts by unit time. Invariance naturally extends to any constant number of unit time shifts.

***Unary Operators***

The identity function, $a = a$, is one of two unary operators. The *increment* function, $a = b + 1$, is the other. The increment function naturally extends to the addition of any constant. In temporal terms, the increment function implements a unit delay.

***2-ary Operators***

Temporal ordering relationships are a natural way of describing the 2-ary[1] operators. Table 1 contains all such functions of two inputs. The three left columns are associated with a set of three disjoint ordering relationships between inputs *a* and *b*: $a < b$, $a = b$, and $b < a$. For a given operation and for each of these three input relationships, there are three possible outputs: *a*, *b*, or ∞. This suggests $3^3 = 27$ total 2-input functions. However, after accounting for symmetries and removing duplicates, there are 10 unique 2-ary operations; all are shown in Table 1.

In all cases, if the stated relation is true, then the output is equal to one of the inputs (the first input, *a*, by convention here); if the relation is false, then the output is ∞. Hence, for example, $a \prec b$ and $b \succ a$ are not the same function.

[1] The term "binary" is avoided because of ambiguity with respect to binary encoded data.

**Table 1. All 2-ary *s-t* functions.**

| $a < b$ | $a = b$ | $b < a$ | function | name | symbol |
|---|---|---|---|---|---|
| $a$ | $a$ or $b$ | $b$ | if $a < b$ then $a$ ; else $b$ | *min* | $\wedge$ |
| $a$ | $a$ or $b$ | $\infty$ | if $a \leq b$ then $a$ ; else $\infty$ | *less or equal* | $\preccurlyeq$ |
| $a$ | $\infty$ | $a$ | if $a \neq b$ then $a$ ; else $\infty$ | *not equal* | $\neq$ |
| $a$ | $\infty$ | $b$ | if $a < b$ then $a$ else if $b < a$ then $b$ ; else $\infty$ | *exclusive min* | x$\wedge$ |
| $a$ | $\infty$ | $\infty$ | if $a < b$ then $a$ ; else $\infty$ | *less than* | $\prec$ |
| $b$ | $a$ or $b$ | $a$ | if $a \geq b$ then $a$ ; else $b$ | *max* | $\vee$ |
| $b$ | $\infty$ | $a$ | if $a > b$ then $a$ else if $b > a$ then $b$ ; else $\infty$ | *exclusive max* | x$\vee$ |
| $\infty$ | $a$ or $b$ | $a$ | if $a \geq b$ then $a$ ; else $\infty$ | *greater or equal* | $\succcurlyeq$ |
| $\infty$ | $a$ or $b$ | $\infty$ | if $a = b$ then $a$ ; else $\infty$ | *equal* | $\equiv$ |
| $\infty$ | $\infty$ | $a$ | if $a > b$ then $a$ ; else $\infty$ | *greater than* | $\succ$ |

Symbols for the primitive functions are shown in Figure 2.

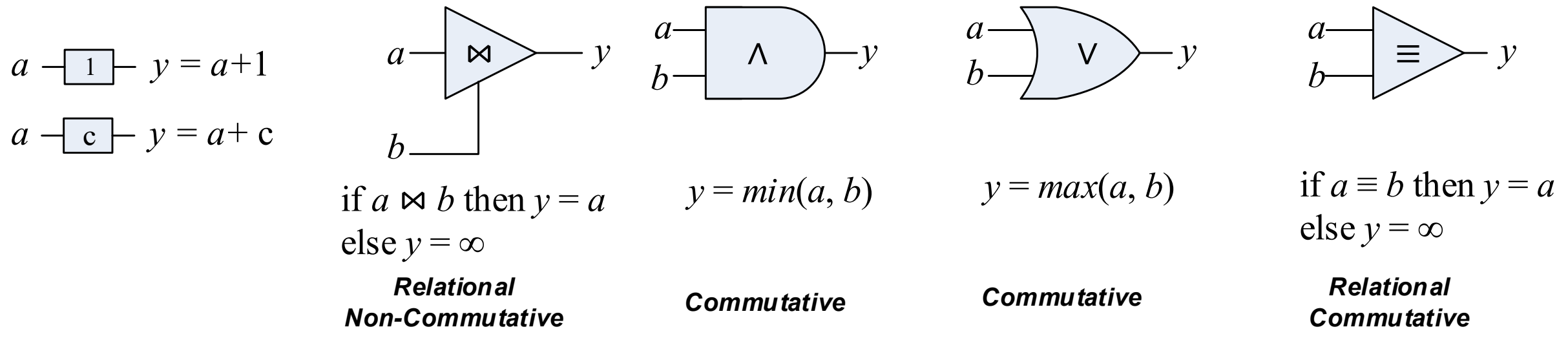

**Figure 2. Symbols representing the various primitive operators that may be used in network schematics. The symbol ⋈ represents any of the non-commutative relational operations.**

In written notation, the asymmetry of inputs is represented by the order of the inputs; that is, $a \bowtie b$ is always interpreted as: "if $a \bowtie b$ then $a$; else $\infty$". The input asymmetry in the drawn relational symbols mnemonically express their non-commutative relationship.

As noted above, the operators $\wedge$ and $\vee$ are commutative, associative, distributive, and satisfy the absorption laws. These, and a number of useful identities are given in [12], and it is shown that the three operators +1, $\wedge$, and $\prec$ are functionally complete for all *s-t* functions.

## 2.2 *s-t* Computing Networks

Networks can be constructed by composing the 1-ary and 2-ary *s-t* operators. This is analogous to the construction of Boolean networks by composing *and*, *or*, and *not* gates.

**Theorem:** Any function implemented as a non-recurrent composition of space-time functions is a space-time function [12].

A *feedforward space-time computing network is* a non-recurrent interconnection of *s-t* functional blocks. Each block represents an implementable, causal, and invariant function.

From Theorem 1, all *s-t* computing networks implement *s-t* functions, so if the designer begins with a set of basic *s-t* operators and the operators are interconnected in any feedforward manner, the overall system must implement an *s-t* function.

*It is important to note that building blocks are not restricted to the basic operators given in Table 1. Any functional block whose input/output behavior satisfies the* s-t *properties can be used.*

## 3. System Architecture

Systems are partitioned into synchronized *segments* that are internally feedforward with feedback between segments (Figure 3). Values are transmitted as waves, or volleys, of spikes. The individual spikes encode values as relative times, and synchronization is provided by a gamma clock that generates a reset spike with a fixed period of *k* time units. It is immediately apparent that communicated values cannot have an infinite range as in the *s-t* algebra described above. Rather finite *s-t* algebras, described in the next subsection, are used.

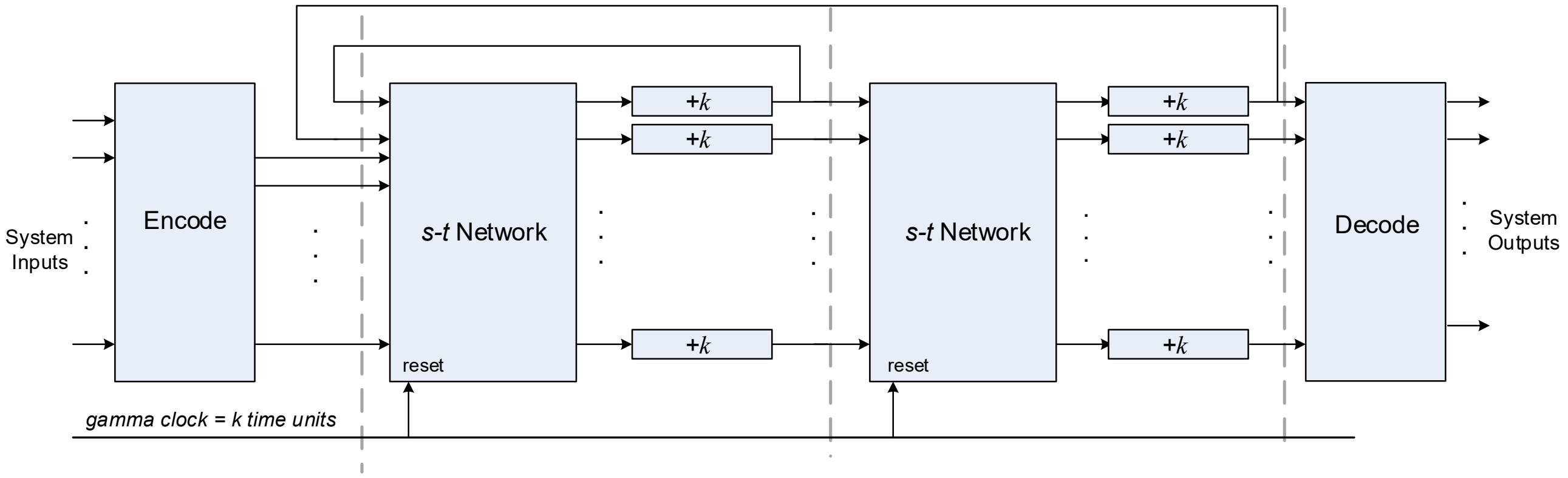

**Figure 3. The system architecture consists of segments; two are shown here. Each segment consists of an *s-t* computing network followed by delays of *k* time units on all segment output lines. Synchronization is provided by a gamma clock with a period of *k* time units. In an implementation, the gamma clock resets, or initializes, computational elements at the beginning of each cycle (see Section 5.1).**

### 3.1 Finite *s-t* Algebras

For practical implementations, the functions implemented by a segment have finite domains and ranges. This suggests finite algebras $\mathbb{S}_k$ defined over a finite set containing $k$+1 elements: $\{0,1,...k\text{-}1, \infty\}$. Finite *s-t* functions satisfy causality and a modified form of temporal invariance:

$$\text{if } F(x_{1,}...x_q) + 1 < k \text{ then } F(x_1 + 1, ..., x_q + 1) = F(x_{1,}...x_q) + 1$$
$$\text{else } F(x_1 + 1, ..., x_q + 1) = \infty .$$

With regard to the primitive operations, the only change is for the increment (delay) function.

$$\text{if } a < k\text{-}1 \text{ then } y = a + 1$$
$$\text{else } y = \infty.$$

To summarize, temporal computing networks are composed of building blocks that implement *s-t* functions based on a finite *k*-valued algebra $\mathbb{S}_k$. In the model all the gates have zero delay except for explicit delay elements. Values are measured in model time units, and the gamma clock has a period of period of *k* time units, i.e., $\gamma = k$ .

### 3.2 Computation Flow

For a given segment, computation proceeds in the following manner as spike volleys flow through.

1) At the beginning of cycle $\gamma_i$ , the *s-t* network is initialized (reset), and all spikes internal to the network are nullified as a byproduct.

2) During cycle $\gamma_i$, an input volley is applied with input values belong to the set $\{0,1,...k\text{-}1,\infty\}$. The *s-t* network computes an output volley.

3) Outputs from the *s-t* network pass through the $+k$ delay network. Because the non-$\infty$ outputs of the *s-t* network range from 0 to $k$-1 during gamma cycle $\gamma_i$, after the delay of $k$, the non-$\infty$ inputs to the subsequent segment range from $k$ to $2k$-1 with respect to $\gamma_i$. Consequently, the inputs to the next segment, referenced with respect to $\gamma_{i+1}$ , belong to the set $\{0,1,...k\text{-}1,\infty\}$.

## 4. Non-*s-t* Segments

Observe that not all functions with domains/ranges belonging to the set $\{0...k\text{-}1, \infty\}$ can be implemented using only *s-t* operators. Importantly, non-causal functions cannot be implemented in the algebra.

### 4.1 Using Gamma as a Temporal Reference

To compute non-*s-t* functions, segments using the gamma clock as a temporal *reference signal* are constructed. Then, given the additional reference input, any function on $\{0...k\text{-}1, \infty\}$can be computed using only *s-t* operations. When used this way, the gamma clock is more than a resetting mechanism, its temporal value becomes an explicit part of the function computation.

As an example, consider the function that reverses values: $Y(A) = k\text{-}1\text{-}A$; if $k=4$ then $Y(0) \rightarrow 3$, $Y(1) \rightarrow 2$, $Y(2) \rightarrow 1$, $Y(3) \rightarrow 0$. An implementation is in Figure 4.

This implementation is laid out in a standard form to be introduced in Section 4.4. Inputs at times 0, 1, 2, and 3 generate outputs at times 7, 6, 5, and 4 *with respect to* $\gamma_i$. With respect to gamma cycle $\gamma_{i+1}$, these are 3, 2, 1, and 0.

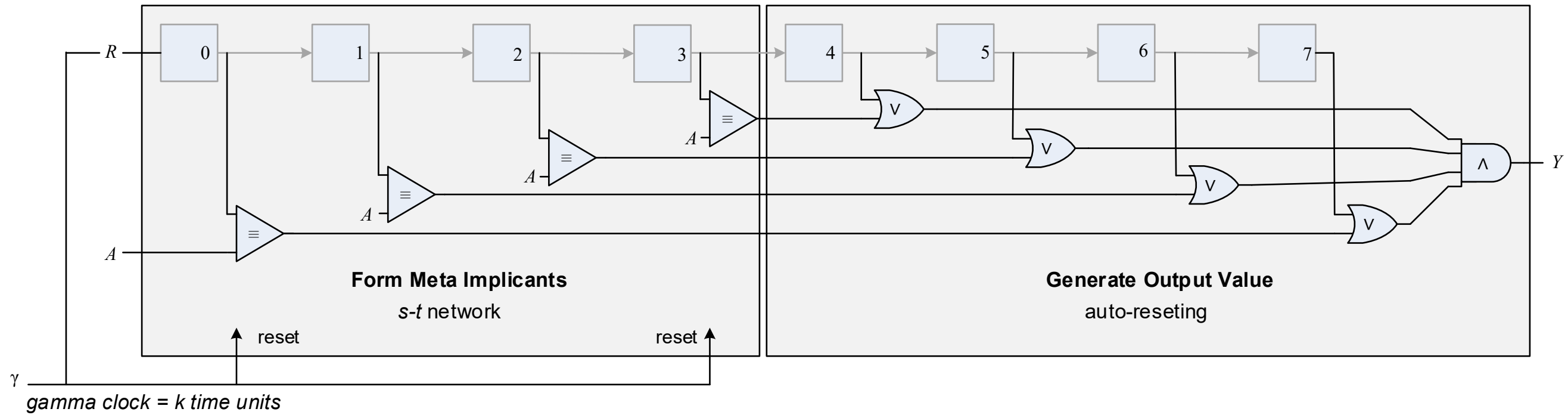

**Figure 4. A non-*s-t* segment that performs the value reversal function. As done throughout this document, the reference spike R is provided by the gamma clock. This is an example of a standard form implementation as described in Section 4.4. The output generation network is auto-resetting because a spike on any of the *max* gates will always be followed by a spike on the delay chain input. I.e., if a meta implicant spike ever sets state in the gate, it will always be reset during normal operation.**

## 4.2 Implementing Arbitrary Functions with Reference Signals

The requirements of causality and invariance restrict the functions that can be implemented with the *s-t* operators. However, with the addition of a temporal reference signal, any non-*s-t* function can be embedded within a larger *s-t* function, thereby alleviating any concerns regarding functional completeness.

In the following, equations are frequently used. Before proceeding, the operator precedence ordering: +1, ⋈, ∧, ∨ , where ⋈ can be any of the predicate operators. When multiple ⋈ operators are given, parenthesis are used. For example, $a+1 < b+2 \wedge c \vee d+3 = (((a+1) < (b+2)) \wedge c) \vee (d+3)$.

## 4.3 Illustrative Example: Quaternary Half Adder

Without loss of generality (due to invariance) define input reference signal R = 0. The immediately following reference signal is denoted as $R_+$ where $k = (R_+ - R)$. In implementations, the gamma clock is used as the reference signal. The sum S′ ranges from 0 to 3 and the $C_{out}'$ signal is 0 or 1. In this example, $k = 4$, so the maximum output value is $3 = k - 1$ and $R_+ = R + 4$. The conventional function table for a quaternary adder is in Figure 5 left, and the table with reference signals added is on the right.

| A′ | B′ | S′ | Cout′ |
|---|---|---|---|
| 0 | 0 | 0 | 0 |
| 0 | 1 | 1 | 0 |
| 0 | 2 | 2 | 0 |
| 0 | 3 | 3 | 0 |
| 1 | 0 | 1 | 0 |
| 1 | 1 | 2 | 0 |
| 1 | 2 | 3 | 0 |
| 1 | 3 | 0 | 1 |
| 2 | 0 | 2 | 0 |
| 2 | 1 | 3 | 0 |
| 2 | 2 | 0 | 1 |
| 2 | 3 | 1 | 1 |
| 3 | 0 | 3 | 0 |
| 3 | 1 | 0 | 1 |
| 3 | 2 | 1 | 1 |
| 3 | 3 | 2 | 1 |

| R reference | | | | | R+ reference | | |
|---|---|---|---|---|---|---|---|
| R | A | B | S | Cout | R+ | S | Cout |
| 0 | 0 | 0 | 4 | 4 | 4 | 0 | 0 |
| 0 | 0 | 1 | 5 | 4 | 4 | 1 | 0 |
| 0 | 0 | 2 | 6 | 4 | 4 | 2 | 0 |
| 0 | 0 | 3 | 7 | 4 | 4 | 3 | 0 |
| 0 | 1 | 0 | 5 | 4 | 4 | 1 | 0 |
| 0 | 1 | 1 | 6 | 4 | 4 | 2 | 0 |
| 0 | 1 | 2 | 7 | 4 | 4 | 3 | 0 |
| 0 | 1 | 3 | 4 | 5 | 4 | 0 | 1 |
| 0 | 2 | 0 | 6 | 4 | 4 | 2 | 0 |
| 0 | 2 | 1 | 7 | 4 | 4 | 3 | 0 |
| 0 | 2 | 2 | 4 | 5 | 4 | 0 | 1 |
| 0 | 2 | 3 | 5 | 5 | 4 | 1 | 1 |
| 0 | 3 | 0 | 7 | 4 | 4 | 3 | 0 |
| 0 | 3 | 1 | 4 | 5 | 4 | 0 | 1 |
| 0 | 3 | 2 | 5 | 5 | 4 | 1 | 1 |
| 0 | 3 | 3 | 6 | 5 | 4 | 2 | 1 |

**Figure 5. Function table for quaternary half adder. The conventional table is on the left; the table with reference signals added is on the right. The implemented *s-t* function satisfies the "R reference" part of the table. When these spikes are re-referenced with respect to $R_+$, the function outputs are shown in the "R+ reference" part of the table; this is what a subsequent segment observes as inputs.**

Equations for the functions *S* and $C_{out}$ can be translated to a minterm standard form. For each row in the table, a *minterm* is generated by expressing input and output values with respect to the reference input R. So, for example, the minterm associated with the first row for *S* is $(A \equiv R \vee B \equiv R) \vee R{+}4$ . Because R equals 0 by definition, this minterm is satisfied if $A = 0$ and $B = 0$. And if both are true, performing the *max* with R+ 4 yields a sum of $S = R + 4 = 4$. (Causality is satisfied because $S \geq A,B$.)

The equation notation can be simplified significantly by using the following shorthand: a number "*n*", without the "+" sign denotes "R + *n*". So, the above example expression is simplified to: $A \equiv 0 \vee B \equiv 0 \vee 4$ . As before, if A and B are both 0, then the output is 4.

## 4.4 Standard Form

There are many ways of implementing the non-*s-t* segments. A particularly useful implementation form is a *standard form*, illustrated in Figure 6. The *s-t* network first forms *meta implicants* for each of the output functions. For a given output function and a given output value *i* for that function, a meta implicant $M_i$ is a subfunction that yields a non-∞ value *i prior to* output generation. In terms of the function table, a meta implicant corresponds to all the rows that evaluate to the same output value *i*. For example, the $C_{out}$ function consists of two meta implicants: $M_4$ that corresponds to all the 4 outputs and $M_5$ that corresponds to all the 5 outputs. These values use the R reference; they become outputs 0 and 1 when the $R_+$ reference is applied at the beginning of the next segment. The half adder S function has 4 meta implicants that evaluate to 4 through 7 using the R reference.

In equation form, a meta implicant is written as $M_i$ = [*s-t equation using delay chain values* 0...*k*-1] ∨ *i*, where $i > k$-1. The overall function is then $F = M_k \wedge M_{k+1} \dots \wedge M_{2k-1}$.

In the half adder example, the equation for *S* is composed of minterms only; there are no further minimizations if we are restricted to a "two-level" standard form. The implicants (minterms) for a few example rows:

$$S = ( A \equiv 0 \vee B \equiv 0 \vee 4 ) \wedge ( A \equiv 0 \vee B \equiv 1 \vee 5 ) \wedge ( A \equiv 0 \vee B \equiv 2 \vee 6 ) \dots \wedge \dots$$

$$( A \equiv 3 \vee B \equiv 3 \vee 6 ).$$

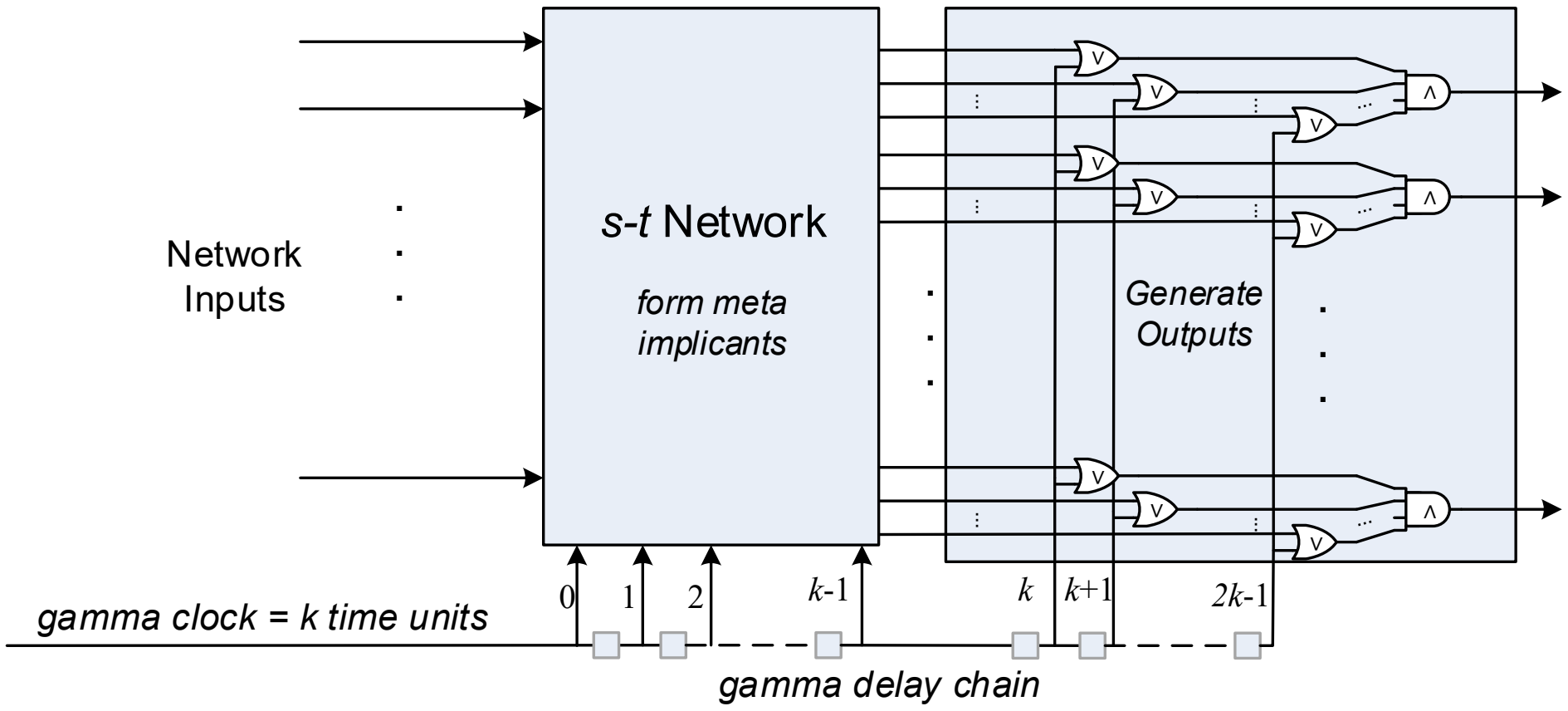

**Figure 6. Standard form non-*s-t* segment. The only delay elements are in the gamma delay chain.**

### 4.5 Optimizations

In the standard form, the only constraint on the *s-t* network is that it is delay free; i.e., all delays are in the gamma delay chain. Otherwise the network can be optimized in any number of ways and operators may be shared during meta implicant formation.

***Implicant Combining***

Implicant combining is an important optimization for individual meta implicants $M_i$. This process is based on Theorem 2 below that allows two implicants to be combined, when 1) all the terms are the same except one, and 2) there is overlap in the terms that differ or they are immediately adjacent. This theorem is essentially a generalization of the Boolean minimization theorem: $(a \wedge b) \vee (a \wedge \sim b) = a$.

Combining can be performed if $(i \preccurlyeq A) \vee (A \preccurlyeq j)$ appears in one implicant and an overlapping (or immediately adjacent) implicant $(n \preccurlyeq A) \vee (A \preccurlyeq p)$ appears in the other. That is, A appears in the interval $(i, j)$ and in the interval $(n, p)$, where $i \leq j$, $n \leq p$. The intervals may overlap (or adjoin) according to the inequalities: $i \leq n$, $n \leq j+1$, $j \leq p$. They adjoin if $n = j+1$. In the following, *T* represents an arbitrary combination of terms that are common to both implicants being combined.

**Theorem 2:**

if $i \leq n,\ n \leq j+1,\ j \leq p$ then

$[(i \preccurlyeq A) \vee (A \preccurlyeq j) \vee T \vee m\,] \wedge [(n \preccurlyeq A) \vee (A \preccurlyeq p) \vee T \vee m\,] = [(i \preccurlyeq A) \vee (A \preccurlyeq p)] \vee T \vee m$

**Pf :**

**lemma 1:** if $a \leq b$ then $(a \preccurlyeq c) \wedge (c \preccurlyeq b) = (a \wedge c)$

**lemma 2:** if $a = b+1$ then $(a \preccurlyeq c) \wedge (c \preccurlyeq b) = (a \wedge c)$

**lemma 3:** if $a \leq b+1$ then $(a \preccurlyeq c) \wedge (c \preccurlyeq b) = (a \wedge c)$ combine lemmas 1 and 2

**lemma 4:** if $a \leq b$ then $(a \preccurlyeq c) \wedge (b \preccurlyeq c) = a \preccurlyeq c$

**lemma 5:** if $a \leq b$ then $(c \preccurlyeq a) \wedge (c \preccurlyeq b) = c \preccurlyeq b$

**lemma 6:** $(a \preccurlyeq b) \vee (a \wedge b) = (a \preccurlyeq b)$

$[(i \preccurlyeq A) \vee (A \preccurlyeq j) \vee T \vee m\,] \wedge [(n \preccurlyeq A) \vee (A \preccurlyeq p) \vee T \vee m\,]$

$= [\,(i \preccurlyeq A) \wedge (n \preccurlyeq A)] \vee [(i \preccurlyeq A) \wedge (A \preccurlyeq p)] \vee [\,(n \preccurlyeq A) \wedge (A \preccurlyeq j)] \vee [(A \preccurlyeq j) \wedge (A \preccurlyeq p)] \vee T \vee m$

apply distributive property twice

$= [(i \preccurlyeq A) \wedge (n \preccurlyeq A)] \vee \boldsymbol{(i \wedge A) \vee (n \wedge A)} \vee [\,(A \preccurlyeq j) \wedge (A \preccurlyeq p)] \vee T \vee m$

$i \leq p$ because $i \leq n$ and $n \leq p$; apply lemma 1

$n \leq j+1$; apply lemma 3

$\boldsymbol{= [\,(i \preccurlyeq A)} \vee (i \wedge A) \vee (n \wedge A) \vee \boldsymbol{(A \preccurlyeq p)]} \vee T \vee m$

$i \leq n$; apply lemma 4

$j \leq p$; apply lemma 5

$= [\,(i \preccurlyeq A) \vee (\,A \preccurlyeq p)] \vee T \vee m$

apply lemma 6 twice

***QED***

Also useful is the following transformation applied at the minterm level:

$[(\boldsymbol{A \equiv i}\,) \vee T \vee m\,] = [(\boldsymbol{A \preccurlyeq i \vee i \preccurlyeq A}) \vee T \vee m\,]$

This follows directly from ***24*** in [12]. In effect, the variable $A$ is expressed as having the same upper bound ( $A \preccurlyeq i$ ) and lower bound ( $i \preccurlyeq A$). Using this identity converts a minterm into the format used in Theorem 2.

***Half Adder Example, contd.***

For the $C_{out}$ function, minterms are combined to form larger implicants. In the example half adder design, consider meta implicant $M_4$. The second and third rows of the function table for $C_{out}$ are minterms that are a subfunction of $M_4$:

$[\,A \equiv 0 \vee B \equiv 1\,] \wedge [\,A \equiv 0 \vee B \equiv 2]$

Using the minimization theorem, these two minterms can be combined on variable $B$ to yield the implicant:

$[\,A \equiv 0 \vee (1 \preccurlyeq B\,) \vee (B \preccurlyeq 2)\,]$.

In fact, the first four rows of $C_{out}$ can be combined to form a single meta implicant:

$$[\, A \equiv 0 \vee (0 \preccurlyeq B)] \wedge [\, A \equiv 0 \vee (B \preccurlyeq 3)]$$

At this point, consider the following identities when $B \neq \infty$ :

$0 = 0 \preccurlyeq B$

$B = B \preccurlyeq k\text{-}1$

$B \vee k = k$

For the example, $k$-1 = 3:

$0 = 0 \preccurlyeq B$

$B = B \preccurlyeq 3$

$B \vee 4 = 4$

Re-write the meta implicant in its complete form and apply the above identities:

$$[\, A \equiv 0 \vee \mathbf{0 \preccurlyeq \boldsymbol{B}} \vee 4\,] \wedge [\, A \equiv 0 \vee \boldsymbol{B} \mathbf{\preccurlyeq 3} \vee 4\,]$$

$$= [\, A \equiv 0 \vee \mathbf{0} \vee 4\,] \wedge [\, A \equiv 0 \vee \boldsymbol{B} \vee 4\,]$$

$$= [\, A \equiv 0 \vee 4\,].$$

In the end, the variable $B$ is eliminated because the full range of B is included in the original meta implicant.

Applying this method repeatedly yields a standard form equation for $C_{out}$ that contains a meta implicant $M_4$ composed of four implicants and a meta implicant $M_5$ that is a constant.

$$C_{out} = [\, A \equiv 0 \vee 4\,] \wedge [\, B \equiv 0 \vee 4\,] \wedge [\, A \equiv 1 \vee B \preccurlyeq 2 \vee 4\,] \wedge [\, B \equiv 1 \vee A \preccurlyeq 2 \vee 4\,] \wedge 5\,.$$

***Equality Combining***

This optimization combines minterms based on temporal invariance and is best explained via an example. Begin with the equation for $S$ composed of minterms.

Consider the function table rows: [A B] = [0 0], [1 1], [2 2], and [3 3]. These produce the minterms:

$$(\, A \equiv 0 \vee B \equiv 0 \vee 4\,) \wedge (\, A \equiv 1 \vee B \equiv 1 \vee 6\,) \wedge (\, A \equiv 2 \vee B \equiv 2 \vee 4\,) \wedge (\, A \equiv 3 \vee B \equiv 3 \vee 6\,)$$

The gate cost is: 8 $\equiv$ , 3 $\wedge$, 8 $\vee$,   19 total.

Applying invariance:

$$= [\, (A \equiv B) \equiv 0 \vee 4\,] \wedge [\, (A \equiv B) \equiv 1 \vee 6\,] \wedge [\, (A \equiv B) \equiv 2 \vee 4\,] \wedge [\, (A \equiv B) \equiv 3 \vee 6\,]$$

$$= [(A \equiv B) \equiv 0 \wedge (A \equiv B) \equiv 2 \vee 4] \wedge [\, (A \equiv B) \equiv 1 \wedge (A \equiv B) \equiv 3 \vee 6]$$

The gate cost is: 5 $\equiv$ ,  3 $\wedge$,  2 $\vee$,  10 total

Observe that the result is in standard form with the operator ($A \equiv B$) being shared between two meta implicants.

## 5. Synchronization

### 5.1 Gamma Synchronization

The gamma clock plays a role similar to inhibitory oscillations in biological neural networks as described by Fries et al. [3]. That is, it performs a state reset operation between waves of computational activity. The model does not explicitly model inhibitory neurons, only the functional effect of inhibitory oscillations. Consequently, synchronization provided by the gamma clock performs the same high level function as the clock in classical synchronous systems: it supports reliable feedback and overlapped computation between stages (pipelining).

In the model, there are two types of state that rely on the gamma clock for resetting: explicit state and implicit state.

***Explicit State***

Delay elements contain explicit state. Because the delay element function is limited by $k$-1 (Section 3.1) delay elements are implemented with a reset signal $r$ provided by the gamma clock which marks off intervals of size $k$ (0 to $k$-1). By definition of the gamma clock, $r$ always occurs at relative time 0. The delay element is implemented as a small pulse mode asynchronous machine with the state transition table given in Figure 7. In effect, reset nullifies pending spikes at the delay element's output.

A system wide assumption is that between gamma spikes, a given line never spikes more than once. Consequently, an "E" in the table indicates cases where this assumption must have been violated. Applying a reset to delay elements every $k$ cycles guarantees that no delay output value can exceed $k$-1.

| *delay (+1)* | | | | |
|---|---|---|---|---|
| current state | input (*ar*) | | next state | output |
| Y0 | - - | | Y0 | - |
| Y0 | τ - | | Y1 | - |
| Y0 | *0 0* | | Y1 | - |
| Y0 | - 0 | | Y0 | - |
| Y1 | - - | | Y0 | τ+1 |
| Y1 | τ - | | E | E |
| Y1 | *0 0* | | Y1 | - |
| Y1 | - 0 | | Y0 | - |

**Figure 7. State transition table for the pulse mode asynchronous delay element. The symbol τ is a value between 0 and *k*-1.**

***Implicit State***

Consider the temporal operation $a \prec b$ : if a spike on $a$ occurs before a spike on $b$ then the output is a spike at the same time as $a$; else it is ∞. It seems that any plausible implementation of this operation will contain at least one technology-dependent bit of state -- to keep track of whether or not spike $b$ ($a$) has already occurred at the time spike $a$ ($b$)arrives. This is an example of *implicit* state (*i*-state). Consequently, in an implementation that continually re-uses the same operator hardware (and any practical implementation would do so), the *i*-state will have to be initialized by resetting it at the very beginning of each gamma cycle.

In some technologies the reset may happen on its own. For example, after being set, the *i*-state may decay back to a reset state by the end of a gamma cycle. In other implementations, reset may be explicit via a gamma cycle reset signal that is broadcast to all the operators (gates) containing *i*-state.

The latter approach leads to *s-t* operators that are implemented as pulse-mode asynchronous machines. (The ≡ and ≠ operators are exceptions). Ideally, the state machine for an operator will be in an initial state at the beginning of each gamma cycle. However, in the absence of a reset, if only one of an operator's inputs receives a spike during a gamma cycle, the state machine may be left in a non-initial state for the next cycle. Consequently, the gamma clock provides an additional reset input to the state machines.

The state transition tables for *min* and *max* functions are given as examples in Figure 8. The gamma clock is the reset signal *r*.

| *min* | | | | *max* | | | |
| --- | --- | --- | --- | --- | --- | --- | --- |
| current state | input (*abr*) | next state | output | current state | input (*abr*) | next state | output |
| Y0 | - - - | Y0 | - | Y0 | - - - | Y0 | - |
| Y0 | *τ* - - | Y1 | *τ* | Y0 | *τ* - - | Y1 | - |
| Y0 | - *τ* - | Y1 | *τ* | Y0 | - *τ* - | Y1 | - |
| Y0 | *τ τ* - | Y0 | *τ* | Y0 | *τ τ* - | Y0 | *τ* |
| Y0 | - - *0* | Y0 | - | Y0 | - - *0* | Y0 | - |
| Y0 | - *0 0* | Y1 | *0* | Y0 | - *0 0* | Y1 | - |
| Y0 | *0* - *0* | Y1 | *0* | Y0 | *0* - *0* | Y1 | - |
| Y0 | *0 0 0* | Y0 | *0* | Y0 | *0 0 0* | Y0 | *0* |
| Y1 | - - - | Y1 | - | Y1 | - - - | Y1 | - |
| Y1 | *τ* - - | Y0 | - | Y1 | *τ* - - | Y0 | *τ* |
| Y1 | - *τ* - | Y0 | - | Y1 | - *τ* - | Y0 | *τ* |
| Y1 | *τ τ* - | E | E | Y1 | *τ τ* - | E | E |
| Y1 | - - *0* | Y0 | - | Y1 | - - *0* | Y0 | - |
| Y1 | - *0 0* | Y0 | - | Y1 | - *0 0* | Y1 | - |
| Y1 | *0* - *0* | Y0 | - | Y1 | *0* - *0* | Y1 | - |
| Y1 | *0 0 0* | Y0 | *τ* | Y1 | *0 0 0* | Y0 | *0* |

**Figure 8. State transition diagrams for pulse mode asynchronous *min* and *max* function implementations. They each have two implicit states, Y0 and Y1. By definition, the reset signal always occurs at time 0. The symbol τ is any value between 0 and *k*-1. When both *a* and *b* are τ, they have the same value. An output value of τ is the same as the input value that triggers it. E values indicate situations that should never occur because of the requirement that no input receives more than one spike during a single gamma cycle.**

### 5.2 Unit Time Synchronization

This document has focused on the development of a temporal computing model in which delay units (+1) are ideal and gates have zero delay. In real physical implementations, neither is true.

In the model, low precision spike times are essential "rounded" to discrete unit times. In a physical implementation where there are variations in physical components and wires, spike times will gradually diverge as spikes pass through gates and delay elements. If nothing is done, spikes that should be assigned the same value may be assigned different values as spikes drift apart. Consequently, some type of spike realignment will be needed in all but short feedforward networks. In model terms, realignment can be accomplished via special purpose *max* gates having a *data_in* input, a *reset* input, and a *date_out* output. The *reset* input is a low skew unit time clock. If a *data_in* spike occurs any time during a unit time cycle, the immediately subsequent *reset* input yields a *data_out* spike. Because the reset signal is low skew, the skew in the *data_out* spikes will also be low, leading to spike realignment.

The placement of realignment *max* functions depends on the specific implementation, at a minimum any feedback loop should contain at least one. In a structured segmented design as described above, unit time realignment at all segment outputs assures that all feedback loops contain realignment because all the segments are feedforward internally.

## 6. Segment Composition

Figure 9 contains block diagrams for both types of segments considered in this document. Observe that they both have two "phases", the first of which is an *s-t* network that needs to be reset every gamma cycle, the second of which needs no reset. To perform realignment a *max* function (*m*) is placed at segment outputs. Both types of segments use the same gamma clock and a low skew unit time clock. The two types of segments can be combined seamlessly in the same system.

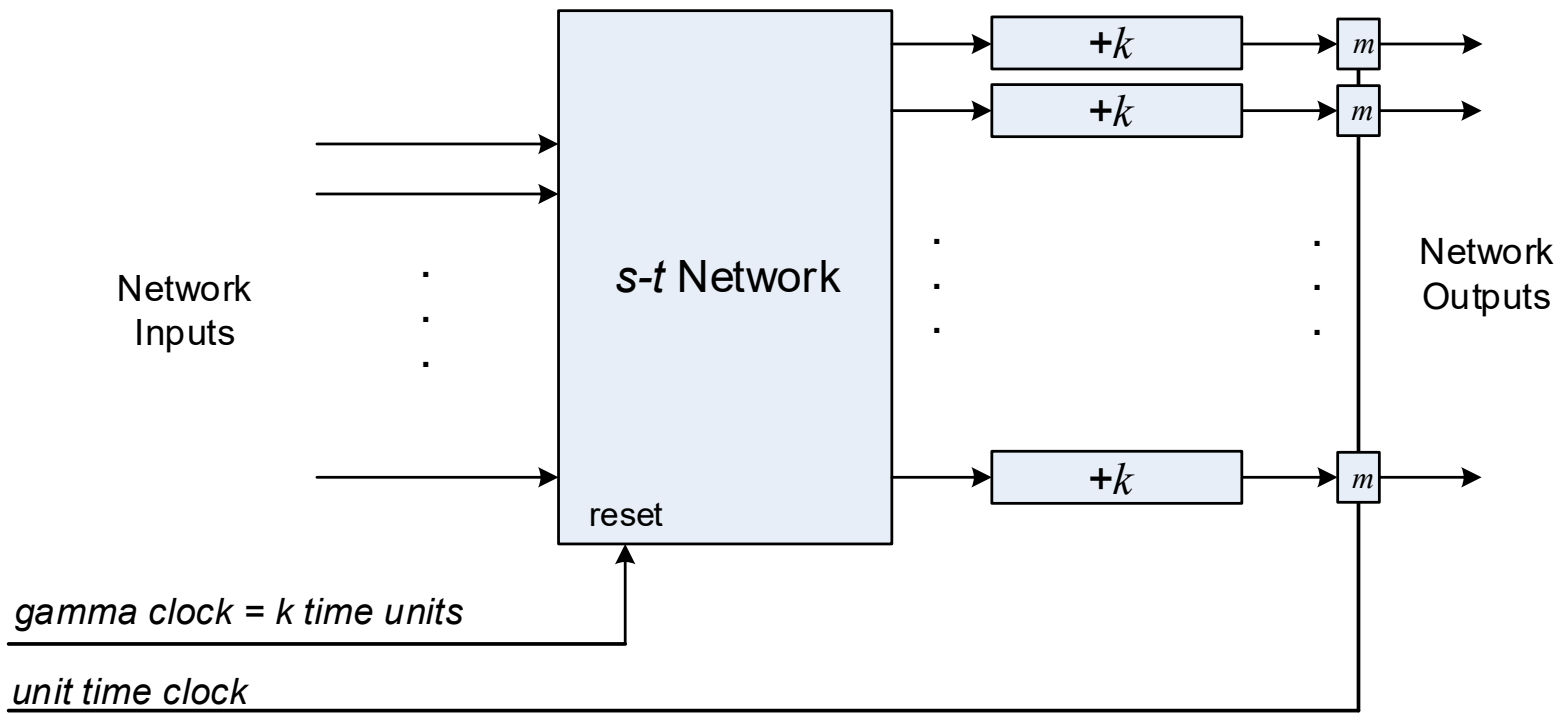

**a) s-t segment**

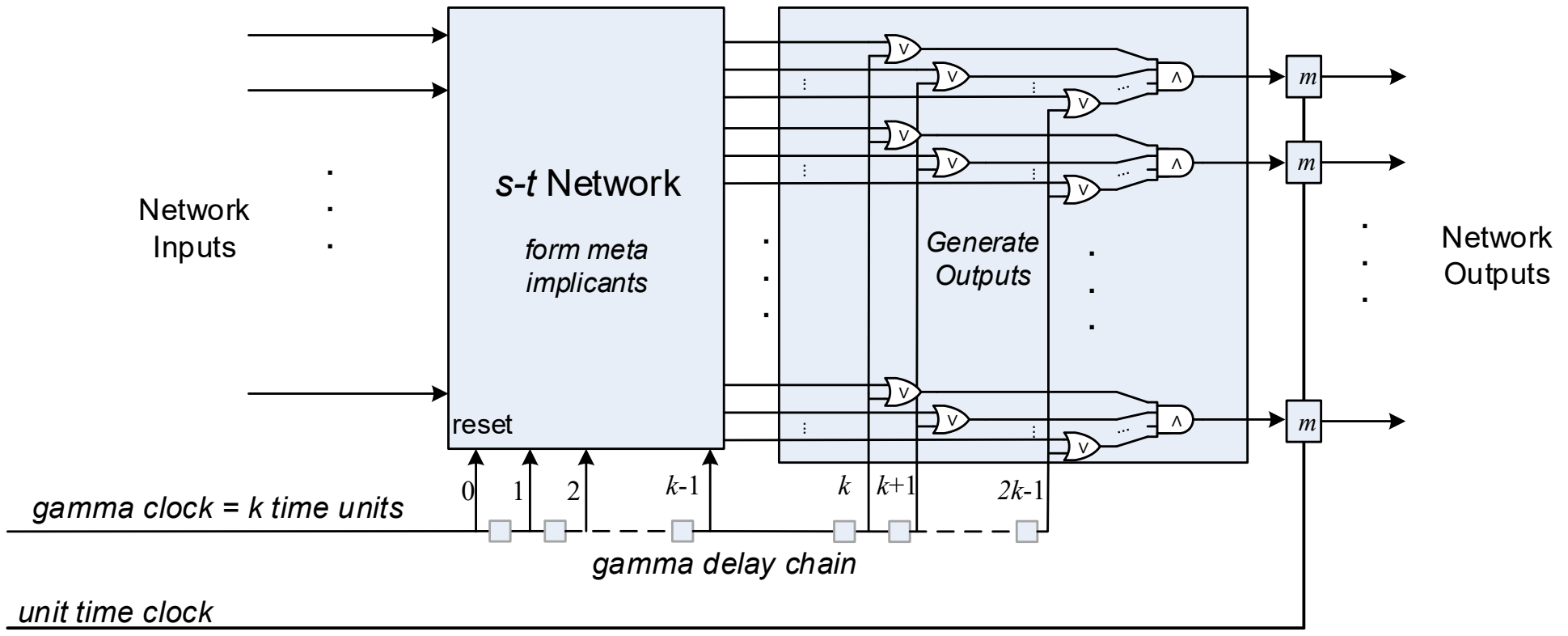

**b) standard form non-s-t segment**

**Figure 9. High level implementations of s-t and non-s-t segments. Both use the gamma clock to reset internal operators and a low skew unit time clock to perform spike realignment. The *m* blocks perform spike realignment.**

Spiking neural networks can be implemented using *s-t* segments as described earlier in this document. If this approach is used then there will likely be a very large number of delay elements. In contrast, the non-*s-t* implementation method has very few delay elements -- they only appear in the gamma delay chain. In fact, a single delay chain could conceivably be shared across an entire system, or at least a large subsystem. Follow up work should consider the implementation of spiking neural networks using non *s-t* segments (even though they meet the constraints for *s-t* segments).

**Acknowledgements:** George Tzimpragos, Tim Sherwood, Harideep Nair, John Shen, Ido Guy, Shlomo Weiss, Rajit Manohar